\documentclass[conference]{IEEEtran}
\IEEEoverridecommandlockouts
\usepackage{cite}
\usepackage{amsmath,amssymb,amsfonts}
\usepackage{algorithmic}
\usepackage{tcolorbox}
\usepackage{multirow}
\usepackage{tabularx}   
\usepackage{graphicx}
\usepackage{textcomp}
\usepackage[normalem]{ulem}
\usepackage[dvipsnames]{xcolor}
\usepackage{url}
\usepackage{subfig}
\usepackage{booktabs}   
\usepackage{xcolor}
\usepackage{soul}
\def\BibTeX{{\rm B\kern-.05em{\sc i\kern-.025em b}\kern-.08em
    T\kern-.1667em\lower.7ex\hbox{E}\kern-.125emX}}
\begin{document}

\title{Beyond Traditional Diagnostics: Transforming Patient-Side Information into Predictive Insights with Knowledge Graphs and Prototypes
}

\author{\IEEEauthorblockN{1\textsuperscript{st} Yibowen Zhao}
\IEEEauthorblockA{\textit{Joint SDU-NTU Centre for} \\
\textit{Artificial Intelligence Research} \\
\textit{\& School of Software} \\
\textit{Shandong University}\\
Jinan, China \\
ybw.zhao@mail.sdu.edu.cn}
\and
\IEEEauthorblockN{2\textsuperscript{nd} Yinan Zhang\IEEEauthorrefmark{1}\thanks{\IEEEauthorrefmark{1} Corresponding authors: Yinan Zhang (yinan.zhang@ntu.edu.sg) and Lizhen Cui (clz@sdu.edu.cn)}
}
\IEEEauthorblockA{\textit{College of Computing and Data Science} \\
\textit{Nanyang Technological University}\\
Singapore \\
yinan.zhang@ntu.edu.sg}
\and
\IEEEauthorblockN{3\textsuperscript{rd} Zhixiang Su}
\IEEEauthorblockA{\textit{College of Computing and Data Science} \\
\textit{Nanyang Technological University}\\
Singapore \\
zhixiang002@e.ntu.edu.sg}
\and
\IEEEauthorblockN{4\textsuperscript{th} Lizhen Cui\IEEEauthorrefmark{1}}
\IEEEauthorblockA{\textit{Joint SDU-NTU Centre for} \\
\textit{Artificial Intelligence Research} \\
\textit{\& School of Software} \\
\textit{Shandong University}\\
Jinan, China \\
clz@sdu.edu.cn}
\and
\IEEEauthorblockN{5\textsuperscript{th} Chunyan Miao
}
\IEEEauthorblockA{
\textit{College of Computing and Data Science} \\
\textit{Nanyang Technological University}\\
Singapore \\
ascymiao@ntu.edu.sg}
}

\maketitle

\begin{abstract}
Predicting diseases solely from patient-side information, such as demographics and self-reported symptoms, has attracted significant research attention due to its potential to enhance patient awareness, facilitate early healthcare engagement, and improve healthcare system efficiency. However, existing approaches encounter critical challenges, including imbalanced disease distributions and a lack of interpretability, resulting in biased or unreliable predictions. To address these issues, we propose the Knowledge graph-enhanced, Prototype-aware, and Interpretable (KPI) framework. KPI systematically integrates structured and trusted medical knowledge into a unified disease knowledge graph, constructs clinically meaningful disease prototypes, and employs contrastive learning to enhance predictive accuracy, which is particularly important for long-tailed diseases. Additionally, KPI utilizes large language models (LLMs) to generate patient-specific, medically relevant explanations, thereby improving interpretability and reliability. Extensive experiments on real-world datasets demonstrate that KPI outperforms state-of-the-art methods in predictive accuracy and provides clinically valid explanations that closely align with patient narratives, highlighting its practical value for patient-centered healthcare delivery.

\end{abstract}


\section{Introduction}
Predicting diseases using only patient-side information, such as demographics and self-reported symptoms, holds significant potential for enhancing patient awareness and healthcare efficiency. This task is clinically motivated and consistent with current practice~\cite{Chamberse027743,Wallace2022}, where patient-reported symptoms are already used in pre-encounter digital triage through tools like digital symptom checkers and virtual triage to guide care and manage workload.
Firstly, many individuals experiencing mild or ambiguous symptoms delay seeking medical care due to uncertainty, cost, or inconvenience. For instance, 37.8\% of adults have reportedly delayed necessary medical care, mainly due to barriers such as cost or low perceived urgency, leading to adverse health outcomes~\cite{ratnapradipa2023factors,taber2015people}. Extended waiting time further exacerbate this issue by discouraging consultation and delaying diagnosis and treatment~\cite{lee2019differences}. A predictive system leveraging patient narratives and demographics can provide preliminary, actionable insights to support timely medical engagement, serving as an assistive triage tool that complements clinical workflows under physician supervision.

The rise of telemedicine and online doctor consultation platforms further highlight the need for accurate patient-side prediction to guide patients through healthcare services. Many patients struggle to identify the right medical specialty, with studies showing that nearly 40\% of emergency visits are non-urgent and could be managed in primary care~\cite{giannouchos2024concordance}. Predictive models that recommend suitable departments or potential disease categories can significantly improve care navigation, allowing more accurate classification and efficient allocation of medical resources.

Moreover, patient-side prediction systems function as valuable clinical decision support tools, particularly in high-volume primary care and outpatient settings~\cite{land2019reassured,rubio2023american}. By structuring symptoms and risk factors into actionable insights, these systems support early clinical assessment, as shown in a Canadian trial where their triage matched physician judgment and reduced patient wait times and confusion~\cite{chan2021performance}.

However, existing approaches to patient-side disease prediction face substantial challenges, as illustrated in Figure 1, notably arising from the imbalanced distribution of disease cases in available datasets. High-prevalence conditions, such as hypertension and type 2 diabetes, dominate medical records, whereaslong-tailed diseases, including rare disorders and non-endemic infections in the dataset’s catchment such as dengue or malaria, are recorded far less often.~\cite{huang2024label,cai2024bpaco,banerjee2023machine}. This imbalance leads to prediction biases and markedly reduced diagnostic accuracy for less prevalent conditions~\cite{ma2019affinitynet,oliveira2024meta}.
Furthermore, most predictive methods rely on shallow textual features without structured medical knowledge~\cite{yu2024multi,CuiM24,su2024enabling}, which often results in missed clinical semantics and disease misclassification.
Moreover, the lack of interpretability remains a significant barrier to applying predictive models in clinical practice. Most existing predictive models do not clearly explain the rationale behind their predictions or clinically justify their outcomes~\cite{chhetri2023towards,zhang2021context}. This lack of transparency undermines clinician confidence, disrupts effective communication between patients and providers, and limits the ability of clinicians to verify diagnostic decisions~\cite{bouazizi2024enhancing,lisboa2023coming}.

\begin{figure}[t] 
\centering 
\includegraphics[width=\linewidth]{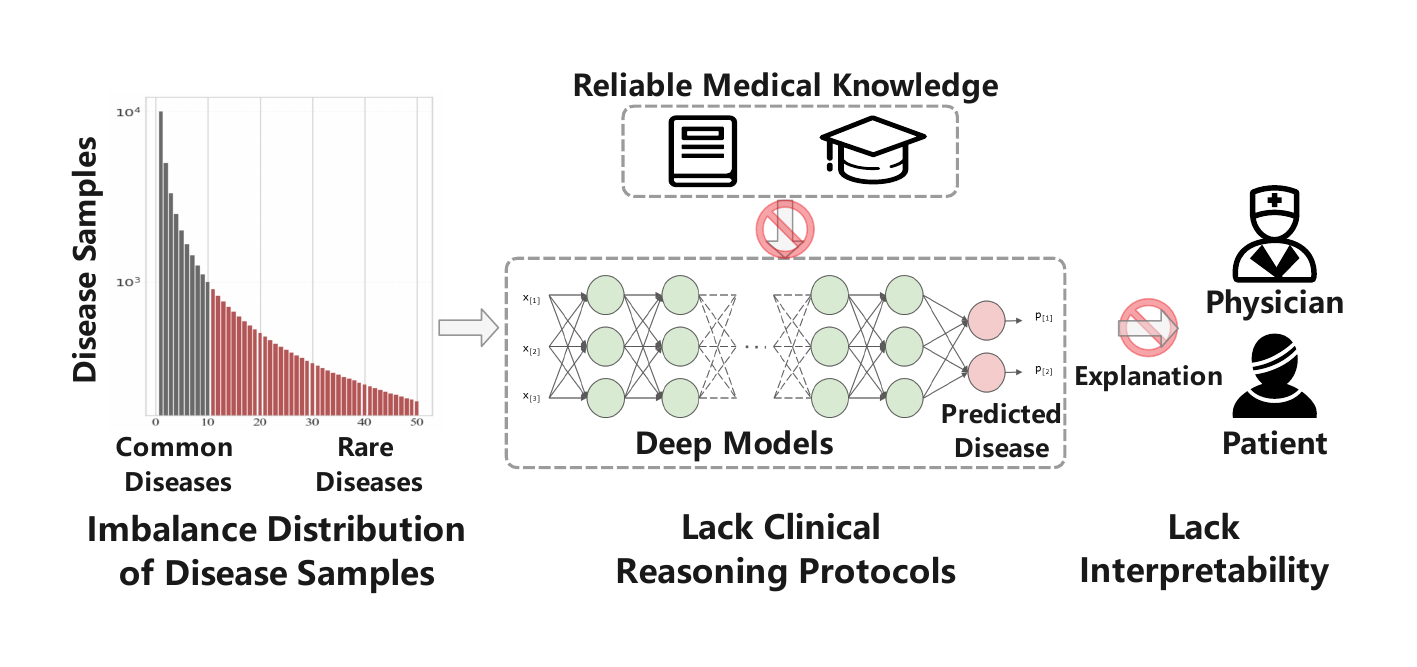} 
\caption{Challenges for Current Patient Disease Prediction.} 
\label{fig:intro} 
\end{figure}

To address these challenges, we propose KPI, a \ul{K}nowledge graph-enhanced, \ul{P}rototype-aware, and \ul{I}nterpretable framework for disease prediction based solely on patient-side information. 
To learn from reliable medical references, KPI constructs a disease knowledge graph that encodes authoritative disease descriptions into a unified KG representation.
KPI then initializes clinically meaningful disease prototypes from the knowledge graph and dynamically refines their embeddings during training, ensuring adaptability to diverse patient narratives while remaining grounded in coherent medical semantics. To accurately classify long-tailed categories, KPI adopts a contrastive learning strategy that aligns patient representations with disease prototypes, thereby enhancing discrimination for underrepresented conditions.
For interpretability, KPI retrieves patient-relevant subgraphs and then leverages large language models (LLMs) to generate coherent, clinically grounded explanations that provide transparent and traceable reasoning.
In summary, the main contributions of this work are as follows:
\begin{itemize}
\item We propose KPI, a novel knowledge-enhanced and interpretable framework for disease prediction from patient narratives. By constructing disease prototypes grounded in structured medical knowledge and generating patient-specific explanations, KPI achieves reliable and clinically valid predictions, even in low-resource or imbalanced disease scenarios.

\item We conduct comprehensive experiments on real-world datasets to validate the effectiveness of KPI. Results show that KPI outperforms state-of-the-art baselines in both predictive accuracy and interpretability. Qualitative case studies further demonstrate that the generated explanations are clinically sound and aligned with patient descriptions.
\end{itemize}

\section{Related Work}


\subsection{Existing Disease Prediction Models}
Existing disease prediction methods mainly focus on structured clinical data, including laboratory results, imaging, and EHRs. 
CNN-based models such as LV-CNN achieve high diagnostic accuracy from images~\cite{jain2023optimized}, while DNN–LightGBM–XGBoost ensembles boost performance on large lab datasets~\cite{park2021development}. Knowledge-graph approaches such as MedPath, SeqCare, and KGxDP integrate biomedical ontologies with EHR sequences for interpretable reasoning~\cite{ye2021medpath,yu2025self,yang2023interpretable}, and GraphCare aligns records with structured knowledge using GNNs~\cite{jianggraphcare}. Explainable AI has further improved trust and uptake, from SHAP-based imaging attribution in coronary disease~\cite{otaki2022clinical} to interpretable EHR models deployed in practice~\cite{lauritsen2020explainable}. However, they rely on high-fidelity, institutionally collected data that are often unavailable at pre-encounter stage.


For conditions that are infrequent in digital records, recent studies address data scarcity and heterogeneity using several strategies. Knowledge graph–assisted models improve classification when only sparse records are available~\cite{li2019improving}; few-shot imaging methods detect rare pathologies with minimal labeled data~\cite{quellec2020automatic}; and large language models extract rare-disease entities and phenotypes from unstructured text~\cite{shyr2024identifying}. Molecular profiling is also promising, with proteomic signatures improving risk prediction across both common and uncommon conditions~\cite{carrasco2024proteomic}. In this work, our focus is the broader problem of imbalanced class distributions that hinder learning, where long-tail categories, including rare diseases, are typically difficult to classify due to limited representation.


In summary, most disease prediction models overlook patient-side information, especially free-text symptom narratives. These narratives provide early, subjective signals that often precede formal evaluation~\cite{hassan2024optimizing}. Ignoring them limits real-world applicability when clinical data is delayed or unavailable. Bridging this gap is crucial for building more responsive, inclusive, and patient-centered predictive systems.

\subsection{Role of Patient Narratives in Diagnosis}
Patient-reported narratives have long been critical to clinical diagnosis, typically collected through history-taking and descriptions of subjective symptoms. Recent research shows that these narratives, whether verbal, written, or digitally expressed, encode diagnostically rich information that is often missing from structured clinical data. For instance, a study demonstrated that a large pre-trained model achieved over 77\% diagnostic accuracy using only patient history text~\cite{fukuzawa2024importance}, highlighting the depth of clinical reasoning embedded in narrative input. Analysis of online rare disease forums has similarly uncovered recurring symptom patterns and diagnostic delays that could serve as early warning signs for clinicians~\cite{al2024quantitative}. Furthermore, PoMP~\cite{su2024enabling} integrates patient narratives with basic demographics to predict diseases, and its strong performance demonstrates the feasibility of patient-side prediction without clinical test results.
Collectively, these findings suggest that patient narratives are critical for early-stage and accurate diagnosis, especially in ambiguous conditions.

Narrative data has also shown practical value in acute and primary care settings. Online forums have been mined to detect early indicators of COVID-19, demonstrating the diagnostic potential of unstructured, user-generated content~\cite{guo2023identifying,li2023text}. Similarly, fuzzy logic–based chatbots have been developed to interpret SMS symptom descriptions, enabling timely illness prediction in low-resource settings where formal clinical evaluation may be unavailable~\cite{omoregbe2020text}.

However, patient narratives remain underutilized in disease prediction. A review of 96 emergency triage models found that only one-third incorporated unstructured narratives, missing critical context beyond vitals and diagnostic codes~\cite{picard2023use}.

\subsection{Large Language Models in Healthcare}
Recent studies underscore the remarkable capabilities of LLMs in processing patient-centered narratives for a variety of clinical tasks. LLMs have demonstrated high-quality performance in clinical summarization, symptom extraction, and decision support. For example, adapted LLMs have been used to generate summaries across diverse modalities, including radiology reports, progress notes, and doctor–patient conversations, with physicians often rating these summaries as equivalent or superior to those written by humans~\cite{van2024adapted}. In another instance, GPT-4 was shown to generate coherent longitudinal summaries of glioblastoma MRI reports, effectively tracking disease progression across multiple visits~\cite{laukamp2024monitoring}. For symptom extraction, GPT-3.5 paired with few-shot prompting and chain-of-thought reasoning outperformed traditional rule-based and UMLS-based approaches across multiple categories of patient-reported symptoms~\cite{khan2025extraction}.

Beyond information extraction and summarization, LLMs have also demonstrated effectiveness in clinical reasoning from unstructured narrative input. GPT-3 achieved a top-3 diagnostic accuracy of 88\% on 48 validated clinical vignettes, significantly outperforming lay users (54\%) and approaching physician-level performance (96\%)~\cite{levine2024diagnostic}. Expanding on this, recent benchmarks show that open-source LLMs such as DeepSeek can match or exceed GPT-4 in diagnosis and treatment recommendation across 125 multi-specialty patient cases~\cite{sandmann2025benchmark}. Furthermore, GPT-4 demonstrated 89\% accuracy in identifying higher-acuity patients across 10,000 emergency department case pairs, which is comparable to the 86\% accuracy achieved by expert clinicians~\cite{williams2024use}. 

These findings highlight the increasing potential of LLMs to accurately interpret free-text clinical narratives across tasks such as summarization, diagnosis, and triage. Their demonstrated alignment with clinical reasoning in real-world settings suggests that LLMs are well-suited for narrative-based diagnostic modeling. Motivated by this evidence, we incorporate LLMs into our framework to improve both the interpretability and clinical relevance of predictions based on patient-reported information.

\begin{figure*}[t] 
\centering 
\includegraphics[width=\textwidth]{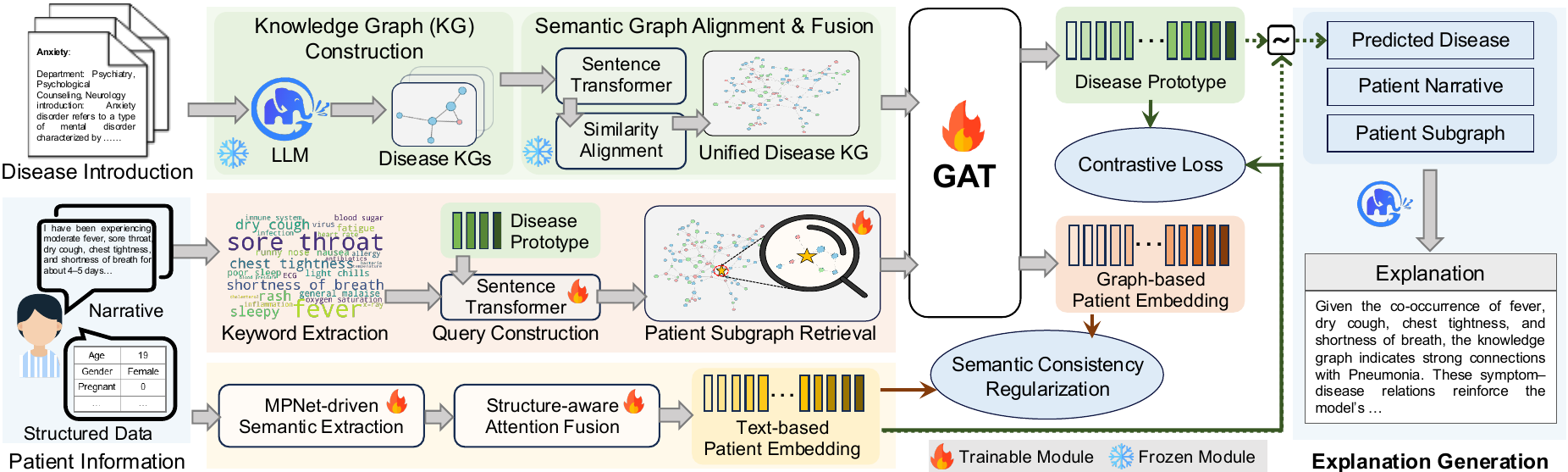} 
\caption{The Architecture of KPI. 
KPI builds a unified disease KG from authoritative descriptions and initializes disease prototypes via graph encoding. For each case, a transformer encodes patient narratives and retrieves a personalized subgraph; contrastive learning aligns the text-based embedding with the correct prototypes, and a semantic consistency regularizer enforces agreement between text- and graph-based patient embedding. At inference (dashed arrows), diseases are predicted by prototype–text similarity, and explanations are generated from the narrative together with its retrieved subgraph.}
\label{fig:framework} 
\end{figure*}

\begin{table}[t]
\centering
\caption{Summary of main notations.}
\label{tab:notation}
\renewcommand{\arraystretch}{1.1}
\small
\resizebox{\linewidth}{!}{
\begin{tabular}{ll}
\toprule
\textbf{Symbol} & \textbf{Description} \\
\midrule
$\mathcal{G}_u$ & Unified disease knowledge graph after alignment and fusion. \\
$\mathcal{G}^{(y)}$ & 2-hop neighborhood around disease $y$ in $\mathcal{G}_u$. \\
$\mathcal{G}_p$ & Patient-specific subgraph. \\
$K_p$ & Set of salient keywords extracted from patient narrative. \\
$P=\{p_j\}$ & Set of disease prototypes. \\
$f(\cdot)$ & Fine-tuned sentence transformer for patient text/keywords. \\
$f'(\cdot)$ & Frozen transformer for KG entity/relation embeddings. \\
$g(\cdot)$ & Graph encoder with attention layers. \\
$q_p$ & Query vector combining patient keywords and disease prototypes. \\
$h_d$ & Embedding of disease prototypes. \\
$h_{gp}$ & Graph-based patient embedding from $\mathcal{G}_p$. \\
$h_{\text{narr}}$ & Patient embedding from narrative. \\
$h_p$ & Text-based patient embedding for prediction. \\
$y$ & Ground-truth disease label. \\
$\hat{y}$ & Predicted disease label. \\
$\alpha_v$ & Attention weight of node $v$ in subgraph retrieval. \\
$\tau$ & Temperature parameter in contrastive loss. \\
$\theta$ & Threshold for keyword selection in TF–IDF. \\
$\delta$ & Similarity threshold for KG entity/relation clustering. \\
$\mathcal{L}_{\text{con}}$ & Prototype-based contrastive loss. \\
$\mathcal{L}_{\text{sem}}$ & Semantic consistency regularization. \\
$\lambda$ & Weight on $\mathcal{L}_{\text{sem}}$ in the final objective. \\
$\mathcal{L}$ & Final training loss: $\mathcal{L}_{\text{con}} + \lambda \cdot \mathcal{L}_{\text{sem}}$. \\
\bottomrule
\end{tabular}
}
\end{table}

\section{Methodology}
As illustrated in Figure~\ref{fig:framework}, KPI is a unified framework for knowledge-enhanced, interpretable disease prediction from patient narratives.
It first builds a disease knowledge graph by parsing authoritative descriptions with an LLM and aligning sources with a sentence-transformer, yielding a unified graph of symptom–disease relations.
For each patient record, it extracts salient terms from the narrative to retrieve a personalized subgraph, while a transformer encodes the narrative and fuses demographics via structure-aware attention to form a text-based patient embedding.
A GAT operates on the unified knowledge graph to obtain disease-prototype embeddings and on the personalized subgraph to produce a graph-based patient embedding.
For model training, a contrastive loss aligns patient embeddings with the correct prototypes, and a semantic consistency regularizer enforces agreement between text- and graph-based patient embeddings.
Finally, KPI predicts the disease through prototype–text similarity and prompts an LLM with the narrative and the retrieved subgraph to generate a concise, clinically grounded explanation.\footnote{The code is available at \url{https://github.com/zybwbnb3/KPI}.}

\subsection{Disease Knowledge Graph Construction}
Generative models, especially large language models (LLMs), often struggle with hallucination by producing confident yet inaccurate or unverifiable outputs. This issue is particularly problematic in clinical settings, where accuracy and trustworthiness are especially essential. Retrieval-augmented generation (RAG) offers a promising solution by grounding model outputs in external, high-quality information~\cite{zhao2025medrag}. The core idea of RAG is to retrieve relevant and reliable content to guide the generation process. Therefore, we are motivated to build a structured and trustworthy medical knowledge graph (KG). A well-constructed KG allows for precise retrieval of clinically validated facts, supports reasoning over disease–symptom relationships, and serves as a robust foundation for downstream tasks such as disease prediction and explanation generation~\cite{gao2025leveraging}.

The construction of a medical knowledge graph, however, presents several key challenges. Firstly, disease descriptions are typically unstructured, with inconsistent terminology, varying levels of detail, and diverse narrative styles. These characteristics make it difficult to accurately extract and standardize medical knowledge. Secondly, LLMs are constrained by input length and often struggle to capture both disease-specific information and general medical semantics simultaneously. 

To address these issues, we propose a two-step framework: (1) \textit{Knowledge Graph Construction via Prompting Pretrained LLMs}, and (2) \textit{Semantic Graph Alignment and Fusion}. Step 1 aims to efficiently structure unstructured medical text using the language understanding capabilities of LLMs. Step 2 ensures consistency across disease graphs by merging and aligning extracted knowledge into a unified, interpretable representation that supports reliable downstream reasoning.

\begin{figure}[t] 
\centering 
\includegraphics[width=\linewidth]{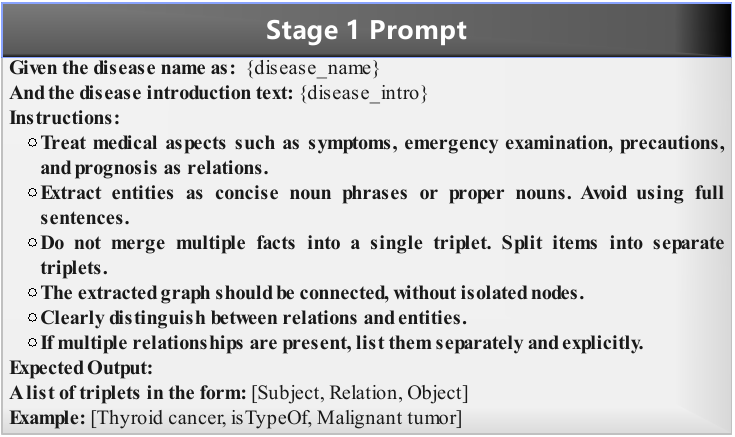} 
\caption{Stage 1: Prompt Template for Extracting Structured Knowledge Triplets from Disease Descriptions.} 
\label{fig:prompt1} 
\end{figure}

\begin{figure}[t] 
\centering 
\includegraphics[width=\linewidth]{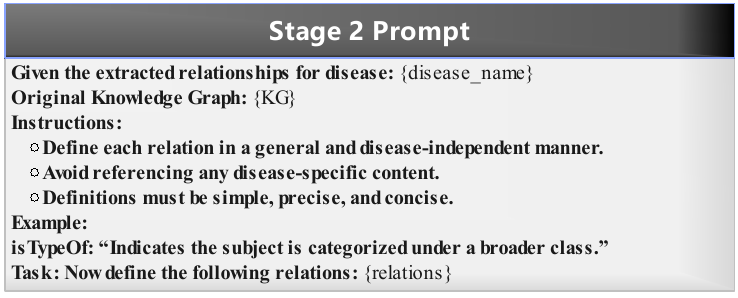} 
\caption{Stage 2: Prompt Template for Defining and Canonicalizing Relation Types Across Diseases.} 
\label{fig:prompt2} 
\end{figure}

\subsubsection{Knowledge Graph Construction via Prompting Pretrained LLMs}
Motivated by the strong natural language understanding and generation capabilities of pre-trained LLMs, we leverage them to transform unstructured disease introduction into structured medical knowledge. This consists of two stages. In the first stage, we design a tailored prompt that guides the LLM to extract medically relevant triplets, which comprise entities ${e_i}$ and their corresponding relations ${r_{ij}}$, from disease introductions obtained from a trusted external source. Entities are defined as concise noun phrases (e.g., `chronic inflammation'), while relations represent clear semantic links (e.g., `causes'). The prompt is shown in Figure~\ref{fig:prompt1}.

However, due to the flexibility of natural language generation in LLMs, the extracted relations from the first stage may include semantically similar expressions with different surface forms (e.g., `leads to' vs. `results in'). This variation makes it difficult to integrate disease knowledge into a unified and coherent graph. To address this, we introduce a second prompting stage that asks the LLM to generalize each extracted relation by removing disease-specific content and providing a concise, disease-independent definition. This step helps unify similar relations, reduces redundancy, and ensures that the resulting knowledge graphs are consistent and interpretable across diseases. The prompt in this stage is shown in Figure~\ref{fig:prompt2}.

As a result, for each disease, we obtain a set of relation triplets accompanied by clear and standardized definitions for each relation type. These definitions serve as a semantic anchor to support downstream alignment, allowing us to merge semantically equivalent relations under a unified label in the subsequent fusion process.

\subsubsection{Semantic Graph Alignment \& Fusion}

Constructing knowledge graphs separately for each disease can lead to redundant entities that represent similar or identical medical concepts, resulting in fragmentation and inconsistency across graphs. To address this, we embed all entities and relations into a shared semantic space using a frozen pre-trained sentence transformer, $f'(\cdot)$, which helps to preserve general biomedical semantics. Specifically, entity embeddings are computed by mean-pooling the token-level embeddings from the transformer:
\begin{equation}
    h_{\text{mean}} = \frac{1}{n} \sum_{i=1}^{n} h_i.
\end{equation}
We then perform semantic alignment and fusion by computing pairwise cosine similarities among all entity and relation embeddings. Items with similarity scores exceeding a predefined threshold $\delta$ are grouped into unified groups, reducing redundancy and improving conceptual coherence across graphs. To ensure interpretability and maintain semantic clarity, each cluster is assigned a canonical label selected as the most frequently occurring surface form among its members (e.g., for a cluster containing leads to, causes, and results in, the label causes would be chosen if it appears most frequently).

This process yields a unified, semantically aligned knowledge graph $\mathcal{G}_u$, where redundancies are minimized and related concepts are coherently grouped. The embeddings of merged entities and relations are initialized by averaging their respective pre-fusion representations, ensuring both semantic consistency and stable initialization for downstream tasks.

\subsection{Patient Subgraph Extraction}
\label{sec:sub_retrival}
A key challenge in patient-side disease prediction is adapting broad medical knowledge to individual patient narratives, which are often sparse, noisy, and vary significantly between individuals. Using the entire unified knowledge graph $\mathcal{G}_u$ directly can introduce irrelevant information and obscure patient-specific symptoms. Thus, it is important to extract patient-specific subgraphs that emphasize only the most relevant medical concepts.

To address this, we design a subgraph extraction module that extracts a personalized subgraph from the unified knowledge graph $\mathcal{G}_u$. This subgraph captures contextual medical knowledge relevant to the patient’s symptoms and history, and serves as the core input for downstream training and interpretability. To retrieve a meaningful subgraph, we first identify the key medical information in the patient's narrative that best represents their condition. Instead of relying on the entire narrative, which may contain irrelevant or ambiguous language, we focus on extracting a concise set of clinically important keywords. These keywords serve as a precise and interpretable query for subgraph retrieval, helping the model concentrate on the most relevant medical concepts while reducing noise from less informative details.

\subsubsection{Keyword Extraction from Patient Narratives}
Free-form patient narratives often include vague descriptions, irrelevant modifiers, and non-clinical context, making it challenging to directly anchor medical reasoning. Simply using the entire narrative may introduce noise and reduce retrieval precision. To mitigate this, we extract a set of salient and medically relevant keywords from each narrative \(T_p\). These keywords serve as interpretable anchors that reflect the core of the patient’s condition. Prior studies have demonstrated that TF–IDF is a computationally efficient and effective method for extracting such clinically meaningful keywords from unstructured medical texts~\cite{chae2023predicting,zhan2021structuring}. We therefore adopt an unsupervised TF-IDF scoring scheme to identify locally distinctive terms that are uncommon across patients but prominent within a given narrative. Formally, let \(\mathcal{P}\) be the set of all patient narratives (with \(\lvert \mathcal{P}\rvert\) total narratives), and let \(V\) denote the vocabulary of unique terms across \(\mathcal{P}\). For a given narrative \(T_p\), let \(c(k, T_p)\) be the raw count of term \(k\) in \(T_p\). We first compute the normalized term frequency:
\begin{equation}
\mathrm{tf}(k, T_p) 
= \frac{c(k, T_p)}{\sum_{k'\in V} c(k', T_p)},
\end{equation}
which captures the relative importance of \(k\) within \(T_p\). Next, we calculate the inverse document frequency as
\begin{equation}
\mathrm{idf}(k)
= \log\!\biggl(\frac{\lvert \mathcal{P}\rvert}{\lvert\{\,T_i\in\mathcal{P}:k\in T_i\}\rvert + 1}\biggr).
\end{equation}
A high IDF value indicates that \(k\) appears in only a few narratives, making it more discriminative.

The TF–IDF score for term \(k\) in \(T_p\) is then defined as the product of these two components:
\begin{equation}
\text{tfidf}(k, T_p)
= \mathrm{tf}(k, T_p)\;\times\;\mathrm{idf}(k).
\end{equation}
Finally, we select the keyword set \(K_p\) by thresholding on this score:
\begin{equation}
K_p = \bigl\{\,k\in V \mid \text{tfidf}(k, T_p) > \theta\bigr\},
\end{equation}
where \(\theta\) is a sparsity-controlling hyperparameter chosen to yield a manageable number of keywords per narrative.

\subsubsection{Prototype-augmented Query Construction}
\label{sec:proto_query}

Even with filtered keywords, aligning them directly with the graph may be suboptimal due to linguistic ambiguity and lack of global context. To further enhance retrieval quality, we incorporate disease-level prior knowledge via prototype embeddings. For each keyword $k_i \in K_p$, we compute an embedding $h_{k_i} = f(k_i)$ using a fine-tuned sentence transformer $f(\cdot)$ adapted for clinical keyword semantics. We further apply a learnable gated aggregation $\rho(\cdot)$ to stabilize keyword representation:
\begin{equation}
\bar{h}_k = \sigma(W_g \bar{h}_{\text{mean}}) \cdot \bar{h}_{\text{mean}} + (1 - \sigma(W_g \bar{h}_{\text{mean}})) \cdot \bar{h}_{\text{max}},
\label{eq:agg}
\end{equation}
where $\bar{h}_{\text{mean}} = \frac{1}{|K_p|}\sum_{k_i \in K_p} h_{k_i}$, $\bar{h}_{\text{max}} = \max_{k_i \in K_p} h_{k_i}$, and $\sigma(\cdot)$ denotes the sigmoid activation.

Let $P = \{p_1, p_2, ..., p_N\}$ denote the set of disease prototypes, with corresponding prototype embeddings $\{h_d^1, h_d^2, ..., h_d^N\}$. We define the mean disease prior as $\overline{h}_d = \frac{1}{|P|} \sum_{p_j \in P} h_d^j$. The query vector $q_p$ is then formed by a convex fusion of prior and evidence:
\begin{equation}
q_p = \beta \cdot \overline{h}_d + (1 - \beta) \cdot \bar{h}_{k},
\end{equation}
where $\beta \in [0,1]$ is a learnable fusion coefficient. This design integrates both local, patient-specific semantics and global, disease-level priors with fewer parameters and improved stability compared with concatenation, and it guides retrieval toward relevant disease neighborhoods while remaining robust to noisy keywords.

\subsubsection{Patient Subgraph Retrieval}
Using hard filtering to select nodes may remove important contextual information and disrupt differentiability during training. On the other hand, including the entire graph can overwhelm the model with irrelevant details. To strike a balance, we adopt a soft attention-based subgraph mechanism that allows the model to focus on the most relevant parts of the graph while preserving useful context.

We first identify the patient’s disease label $y$ (available during training or explanation stages), and retrieve the 2-hop neighborhood subgraph $\mathcal{G}^{(y)}$ around the corresponding disease node in $\mathcal{G}_u$. This constraint reduces computational cost and narrows the focus to clinically proximal information.


We then use a multi-head cross-attention mechanism to compute the relevance between the query vector $q_p$ (as defined in Sec.~\ref{sec:proto_query}) and each node $v$ in the subgraph. For clarity, we write the single-head form; node-wise scores are
\begin{align}
s_v &= \frac{1}{\sqrt{d}} \big(W_q q_p\big)^\top \big(W_k h_v\big), \\
\alpha_v &= \operatorname{softmax}_v(s_v),
\end{align}
where $W_q$ and $W_k$ are projection matrices and $\operatorname{softmax}_v$ normalizes over the nodes of $\mathcal{G}^{(y)}$. The weights $\{\alpha_v\}$ act as patient-conditioned importance scores for entities, inducing a smooth reweighting of node representations while preserving topology and yielding a patient-tailored subgraph representation $\mathcal{G}_p$ that highlights clinically pertinent regions. These representations and importance scores are then used for downstream graph-based patient embedding and explanation. Notably, this subgraph extraction is only performed during training, where the ground truth label $y$ is known, or during explanation, where the predicted label $\hat{y}$ is used.

\subsection{Graph-based Semantic Encoding}
\label{sec:graph_encoding}
Effectively encoding the semantic dependencies and structural nuances in clinical knowledge graphs remains a key challenge, especially given their heterogeneous and often sparse nature. Patient subgraphs, in particular, differ widely in node composition and connectivity due to variations in individual medical histories and data completeness. These complexities underscore the need for an encoding strategy that can adapt to diverse graph structures, preserve medically meaningful context, and support more accurate downstream clinical predictions and decision support.

To address this challenge, we propose a neural graph encoding module that learns contextualized representations across diverse graph structures. Our goal is to preserve both the relational topology and the semantic relevance of medical concepts as they appear in global and patient-specific contexts. To extract meaningful and structurally informed representations from the unified knowledge graph $\mathcal{G}_u$ and the patient-specific subgraph $\mathcal{G}_p$, we introduce a semantic encoding component based on graph attention layers. This component serves as a bridge between graph construction and prototype-based inference, enabling the model to capture the contextual nuances of disease concepts and their interactions. More specifically, we adopt graph attention layers to dynamically weigh neighboring nodes by semantic relevance, which is crucial for handling the heterogeneous, sparse, and personalized nature of our graphs. This approach balances interpretability and efficiency while focusing on the most informative nodes.

Formally, we define the semantic encoding function $g(\cdot)$ as consisting of two stacked graph attention layers and a nonlinear projection head. For a node $v_i$ with initial embedding $x_i$, the encoding process is described as follows:
\begin{align}
    x_i^{(1)} &= \phi\left(\sum_{j \in \mathcal{N}(i)} \alpha_{ij}^{(1)} W^{(1)} x_j \right), \\
    x_i^{(2)} &= \sum_{j \in \mathcal{N}(i)} \alpha_{ij}^{(2)} W^{(2)} x_j^{(1)}, \\
    x_i^{\text{proj}} &= W_o^{(2)}\phi\left(W_o^{(1)} x_i^{(2)}\right),
\end{align}
where $\phi(\cdot)$ denotes the activation function, $\alpha_{ij}^{(\cdot)}$ denotes content-driven attention weights, and $W^{(\cdot)}$, $W_o^{(\cdot)}$ denote trainable parameters.

By applying $g(\cdot)$ to the unified knowledge graph $\mathcal{G}_u$, we obtain enriched node embeddings that encapsulate both intrinsic semantic information and relational context. The embeddings corresponding to disease nodes in $\mathcal{G}_u$ constitute our disease prototypes' embeddings $\{h_d\}$, serving as stable semantic references for subsequent patient matching and prediction tasks.

Similarly, applying $g(\cdot)$ to the patient-specific subgraph $\mathcal{G}_p$ generates structured representations tailored to individual patient contexts. Initial embeddings for nodes within $\mathcal{G}_p$ directly inherit embeddings from their counterparts in the unified graph, ensuring semantic coherence. The resultant node embeddings obtained via $g(\cdot)$ are contextually informed and structured, with the embedding of the central disease node designated as the graph-based patient embedding $h_{gp}$. This embedding facilitates effective downstream alignment and model supervision, reinforcing the interpretability and robustness of patient-disease modeling.

\subsection{Patient Embeddings}
In addition to structured clinical data, patient narratives often contain rich contextual information such as lifestyle factors, symptom descriptions, and psychosocial details, which may be critical for accurate disease prediction~\cite{chen2020emergency,chen2023clinical}. To capture these non-clinical yet potentially important signals, we encode patient narratives into embeddings. This enhances the model's ability to form a more holistic understanding of the patient’s condition, complementing the medical KG-based representation.

More specifically, we employ the sentence transformer $f(\cdot)$ which was also used in Section~\ref{sec:proto_query}. During knowledge graph construction, the transformer remained frozen to preserve general biomedical semantics. However, in subgraph extraction and patient embedding, it is fine-tuned to better capture patient-specific linguistic nuances. Concretely, we first apply $f(\cdot)$ to the narrative $T_p$ to obtain sentence embeddings $\{h_s\}$, and then apply the same gated aggregation $\rho(\cdot)$ as in Eq.~\eqref{eq:agg} to produce the narrative embedding $h_{\text{narr}}$. Additionally, structured demographic information, such as gender and age, is embedded into learnable vectors, $h_g$ and $h_a$, respectively. These vectors are then concatenated with a normalized clinical profile vector $h_c$, forming the structured input $h_s = [h_g; h_a; h_c]$. This combined representation is further refined using multi-head self-attention:
\begin{equation}
    \tilde{h}_s = \psi(\zeta(h_s)),
\end{equation}
where $\zeta(\cdot)$ denotes multi-head attention, and $\psi(\cdot)$ denotes layer normalization. Notably, all structured inputs are self-reported and require no access to clinical records, enhancing the framework’s applicability in patient-side settings.

To integrate textual and structured features, we concatenate the refined structured embedding $\tilde{h}_s$ with the text-based embedding $h_{\text{narr}}$, applying an additional attention mechanism to capture intricate interactions:
\begin{equation}
    \hat{h}_p = \psi(\zeta([\tilde{h}_s; h_{\text{narr}}])).
\end{equation}
The $\hat{h}_p$ is projected into a common semantic space: $h_p = W_o \hat{h}_p$,
where $h_p$ denotes the final text-based patient embedding employed for subsequent disease prediction task.

\subsection{Model Training}
Effective patient-side disease prediction faces two main challenges: imbalanced class distributions and limited structured annotations. These issues hinder generalization to underrepresented classes and weaken alignment with medical knowledge. Models trained solely on free text often struggle with interpretability and may yield unreliable predictions.

To address these challenges, we introduce a prototype-guided framework that fuses structured medical knowledge with multi-form textual patient inputs (free-text narratives and structured textual fields). Knowledge-derived disease prototypes are initialized from the knowledge graph. A contrastive objective aligns each patient embedding with its corresponding prototype, improving class separation under imbalance. A consistency regularizer enforces agreement between embeddings derived from multi-form textual patient inputs and from patient-relevant subgraphs, strengthening semantic grounding. Anchoring predictions to interpretable prototypes yields more robust and trustworthy performance in imbalanced real-world datasets.



\begin{figure}[t] 
\centering 
\includegraphics[width=\linewidth]{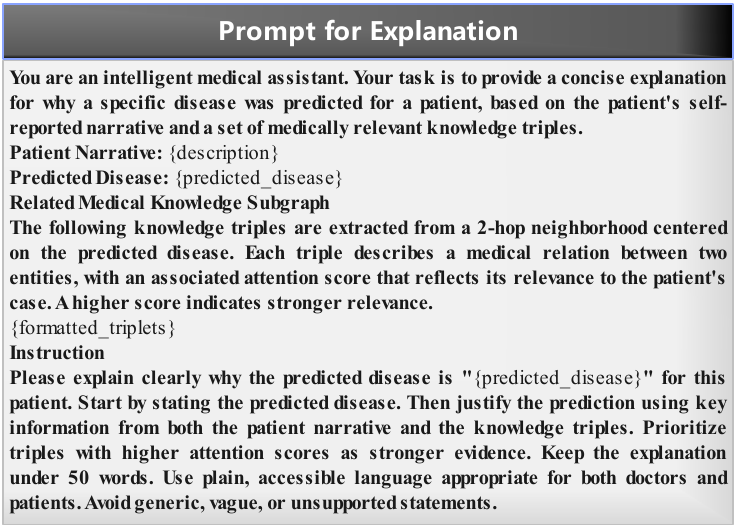} 
\caption{Prompt Template for Generating Patient-Specific Explanations Using Retrieved Subgraphs and Self-Reported Narratives.} 
\label{fig:prompt3} 
\end{figure}

\begin{table}[t]
\caption{Dataset statistics across train/validation/test splits. Patient records denote the number of records per category, and the Pos:Neg ratio is computed in a one-vs-rest setting.}
\centering
\resizebox{\linewidth}{!}{
\begin{tabular}{lcccccc}
\toprule
 & \textbf{Lung} & \textbf{Pneu.} & \textbf{Depr.} & \textbf{CHD} & \textbf{Diab.} & \textbf{Cold} \\
\midrule
Disease Tags     & 55    & 29    & 31    & 63    & 41    & 29    \\ \midrule
Train Records    & 7619  & 4916  & 3979  & 3622  & 3337  & 1118  \\ 
Valid Records    & 979   & 623   & 491   & 464   & 375   & 141   \\
Test Records     & 920   & 604   & 495   & 457   & 445   & 154   \\ \midrule
Total Records    & 9518  & 6143  & 4965  & 4543  & 4157  & 1413  \\ 
Avg. Token       & 188.93 & 208.41 & 241.20 & 197.89 & 176.08 & 196.96 \\
Pos:Neg Ratio    & 1:2.23 & 1:4.00 & 1:5.19 & 1:5.77 & 1:6.39 & 1:20.80 \\
\bottomrule
\end{tabular}
}
\label{tab:sta}
\end{table}

\begin{table*}[]
\caption{Performance Comparison. The best results are bold, and the second-best are underlined.}
\label{tab:main_results}
\resizebox{\textwidth}{!}{
\begin{tabular}{cccccccccccccccccc}
\toprule
Category & Metric & Decision Tree & KNN & AdaBoost & Electra & Finbert & BERT & Albert & Granite & BioLORD & RoBERTa & BioBERT & MedBERT & GTE & GIST & PoMP & KPI \\ \midrule
\multirow{5}{*}{Lung} & hit@1 & 0.1221 & 0.2040 & 0.3238 & 0.3688 & 0.4778 & 0.4845 & 0.5093 & 0.5095 & 0.5171 & 0.5256 & 0.5276 & 0.5470 & 0.5588 & 0.5671 & \underline{0.6101} & \textbf{0.9238} \\
 & hit@3 & 0.2442 & 0.3198 & 0.4998 & 0.3699 & 0.4946 & 0.5095 & 0.5259 & 0.5562 & 0.5645 & 0.5452 & 0.5482 & 0.5688 & 0.5774 & 0.5997 & \underline{0.6596} & \textbf{0.9469} \\
 & hit@10 & 0.2651 & 0.3493 & 0.6550 & 0.4050 & 0.5298 & 0.5520 & 0.5526 & 0.6039 & 0.6016 & 0.5639 & 0.5704 & 0.5957 & 0.6190 & 0.6370 & \underline{0.7722} & \textbf{0.9587} \\
 & AUC & 0.1867 & 0.2292 & 0.4440 & 0.3842 & 0.4984 & 0.5099 & 0.5285 & 0.5743 & 0.5523 & 0.5443 & 0.5477 & 0.5681 & 0.5804 & 0.5948 & \underline{0.6597} & \textbf{0.9374} \\
 & NDCG & 0.3009 & 0.3702 & 0.5539 & 0.4756 & 0.5748 & 0.5852 & 0.6000 & 0.7040 & 0.6224 & 0.6127 & 0.6170 & 0.6344 & 0.6450 & 0.6591 & \underline{0.7263} & \textbf{0.9491} \\ \midrule
\multirow{5}{*}{Pneu.} & hit@1 & 0.0862 & 0.0822 & 0.0496 & 0.0303 & 0.1551 & 0.1791 & 0.1610 & 0.1231 & 0.2007 & 0.1751 & 0.2007 & 0.2230 & 0.2228 & 0.1961 & \underline{0.3081} & \textbf{0.9269} \\
 & hit@3 & 0.2386 & 0.2167 & \underline{0.3368} & 0.0580 & 0.1677 & 0.1923 & 0.1759 & 0.1354 & 0.2190 & 0.1860 & 0.2177 & 0.2402 & 0.2471 & 0.2173 & 0.3320 & \textbf{0.9438} \\
 & hit@10 & 0.5891 & 0.5692 & \underline{0.5934} & 0.0965 & 0.1883 & 0.2140 & 0.1977 & 0.1687 & 0.2381 & 0.2167 & 0.2457 & 0.2573 & 0.2677 & 0.2429 & 0.3446 & \textbf{0.9512} \\
 & AUC & 0.1284 & 0.1245 & 0.2355 & 0.0568 & 0.1713 & 0.1951 & 0.1781 & 0.1904 & 0.2177 & 0.1919 & 0.2188 & 0.2397 & 0.2427 & 0.2163 & \underline{0.3327} & \textbf{0.9370} \\
 & NDCG & 0.3486 & 0.3420 & 0.3931 & 0.1869 & 0.2827 & 0.3032 & 0.2885 & 0.3464 & 0.3222 & 0.3003 & 0.3237 & 0.3409 & 0.3440 & 0.3217 & \underline{0.4612} & \textbf{0.9479} \\ \midrule
\multirow{5}{*}{Depr.} & hit@1 & 0.2097 & 0.2198 & 0.1190 & 0.2086 & 0.3302 & 0.3328 & 0.3463 & 0.2583 & 0.3340 & 0.4672 & 0.3476 & 0.5275 & 0.4806 & 0.4250 & \underline{0.7738} & \textbf{0.8119} \\
 & hit@3 & 0.2367 & 0.3097 & 0.2718 & 0.2510 & 0.3600 & 0.3932 & 0.3706 & 0.2713 & 0.4044 & 0.5328 & 0.3913 & 0.5968 & 0.5406 & 0.5007 & \underline{0.8363} & \textbf{0.8831} \\
 & hit@10 & 0.2387 & 0.3403 & 0.5746 & 0.3057 & 0.4112 & 0.4927 & 0.4170 & 0.3582 & 0.4474 & 0.6237 & 0.4486 & 0.6573 & 0.5736 & 0.5421 & \underline{0.8973} & \textbf{0.9107} \\
 & AUC & 0.2256 & 0.2624 & 0.2535 & 0.2466 & 0.3649 & 0.3867 & 0.3750 & 0.3359 & 0.3778 & 0.5220 & 0.3877 & 0.5781 & 0.5199 & 0.4732 & \underline{0.8175} & \textbf{0.8525} \\
 & NDCG & 0.3396 & 0.3816 & 0.4047 & 0.3621 & 0.4670 & 0.4890 & 0.4727 & 0.5931 & 0.4759 & 0.6032 & 0.4854 & 0.6502 & 0.5960 & 0.5576 & \underline{0.8561} & \textbf{0.8812} \\ \midrule
\multirow{5}{*}{CHD} & hit@1 & 0.1525 & 0.1423 & 0.0299 & 0.1121 & 0.2157 & 0.1525 & 0.1783 & 0.1673 & 0.2658 & 0.1705 & 0.2592 & 0.3416 & 0.2161 & 0.2588 & \underline{0.5549} & \textbf{0.6755} \\
 & hit@3 & 0.1613 & 0.1947 & 0.0907 & 0.1958 & 0.2355 & 0.1907 & 0.2063 & 0.1975 & 0.2890 & 0.2027 & 0.2951 & 0.3667 & 0.2332 & 0.2851 & \underline{0.6239} & \textbf{0.7168} \\
 & hit@10 & 0.1648 & 0.2172 & 0.2401 & 0.3003 & 0.2957 & 0.2620 & 0.2778 & 0.2636 & 0.3454 & 0.2824 & 0.3993 & 0.4371 & 0.2929 & 0.3692 & \underline{0.6995} & \textbf{0.7506} \\
 & AUC & 0.1603 & 0.1714 & 0.1037 & 0.1771 & 0.2466 & 0.1928 & 0.2165 & 0.2558 & 0.2969 & 0.2108 & 0.3046 & 0.3765 & 0.2459 & 0.2982 & \underline{0.6106} & \textbf{0.7054} \\
 & NDCG & 0.3221 & 0.3320 & 0.2626 & 0.3092 & 0.3622 & 0.3192 & 0.3396 & 0.5523 & 0.4065 & 0.3348 & 0.4172 & 0.4771 & 0.3631 & 0.4110 & \underline{0.6854} & \textbf{0.7546} \\ \midrule
\multirow{5}{*}{Diab.} & hit@1 & 0.1342 & 0.1433 & 0.0033 & 0.1469 & 0.4635 & 0.4583 & 0.4019 & 0.3883 & 0.5365 & 0.4035 & 0.4155 & 0.5119 & 0.5135 & 0.4228 & \underline{0.7490} & \textbf{0.7969} \\
 & hit@3 & 0.1490 & 0.2025 & 0.1151 & 0.2107 & 0.4727 & 0.4699 & 0.4209 & 0.4293 & 0.5551 & 0.4166 & 0.4288 & 0.5250 & 0.5309 & 0.4394 & \underline{0.7863} & \textbf{0.8284} \\
 & hit@10 & 0.1504 & 0.2297 & 0.4657 & 0.2711 & 0.5649 & 0.5516 & 0.5065 & 0.4974 & 0.5907 & 0.4835 & 0.4775 & 0.5626 & 0.5806 & 0.5343 & \underline{0.8168} & \textbf{0.8513} \\
 & AUC & 0.1449 & 0.1708 & 0.1254 & 0.2018 & 0.4932 & 0.4862 & 0.4344 & 0.4575 & 0.5591 & 0.4304 & 0.4375 & 0.5327 & 0.5361 & 0.4586 & \underline{0.7787} & \textbf{0.8181} \\
 & NDCG & 0.2626 & 0.2956 & 0.2964 & 0.3299 & 0.5752 & 0.5692 & 0.5256 & 0.5897 & 0.6292 & 0.5201 & 0.5247 & 0.6058 & 0.6070 & 0.5481 & \underline{0.8212} & \textbf{0.8488} \\ \midrule
\multirow{5}{*}{Cold} & hit@1 & 0.0561 & 0.0365 & 0.0000 & 0.0000 & 0.0266 & 0.0227 & 0.0138 & 0.0012 & 0.0395 & 0.0149 & 0.0261 & 0.0679 & 0.0618 & 0.0324 & \underline{0.1039} & \textbf{0.8808} \\
 & hit@3 & 0.0646 & 0.0814 & 0.0000 & 0.0000 & 0.0266 & 0.0240 & 0.0163 & 0.0012 & 0.0432 & 0.0175 & 0.0351 & 0.0705 & 0.0618 & 0.0324 & \underline{0.1137} & \textbf{0.8979} \\
 & hit@10 & 0.0646 & 0.0884 & 0.0282 & 0.0000 & 0.0320 & 0.0397 & 0.0192 & 0.0026 & 0.0484 & 0.0226 & 0.0578 & 0.0768 & 0.0631 & 0.0456 & \underline{0.1206} & \textbf{0.9186} \\
 & AUC & 0.0641 & 0.0559 & 0.0340 & 0.0047 & 0.0324 & 0.0313 & 0.0206 & 0.0560 & 0.0482 & 0.0224 & 0.0395 & 0.0754 & 0.0674 & 0.0406 & \underline{0.1236} & \textbf{0.8932} \\
 & NDCG & 0.2089 & 0.2097 & 0.1945 & 0.1293 & 0.1550 & 0.1553 & 0.1444 & 0.2715 & 0.1704 & 0.1463 & 0.1644 & 0.1932 & 0.1859 & 0.1638 & \underline{0.2914} & \textbf{0.9120} \\ \bottomrule
\end{tabular}
}
\end{table*}

\subsubsection{Prototype-based Contrastive Loss} To support disease prediction, we treat each disease prototype $h_d^+$ as a class anchor and optimize the text-based patient embedding $h_p$ to be closer to its ground-truth prototype while being contrasted against all others. This design leverages structured prior knowledge to define inter-class semantics and provides stable supervision signals, particularly beneficial in low-resource scenarios. The objective is defined via the InfoNCE formulation:
\begin{equation}
\mathcal{L}_{\text{con}} = - \log \frac{\exp(\cos(h_p, h_d^+)/\tau)}{\sum_{j} \exp(\cos(h_p, h_d^j)/\tau)},
\end{equation}
where $\tau$ denotes a temperature parameter and $\cos(\cdot)$ denotes cosine similarity.

\subsubsection{Semantic Consistency Regularization} While contrastive learning encourages inter-class separability, it does not guarantee that the learned $h_p$ aligns with the structure-informed signal encoded in the graph-based patient embedding $h_{gp}$. To bridge this semantic gap, we introduce a regularization term:
\begin{equation}
\mathcal{L}_{\text{sem}} = \lVert h_{gp} - h_p \rVert_2^2,
\end{equation}
which aligns the text-based embedding $h_p$ with its structure-aware counterpart $h_{gp}$, thereby encouraging the text encoder to implicitly capture relational patterns from the knowledge graph. This proves especially valuable for underrepresented diseases, where the structure-enriched $h_{gp}$ provides strong inductive bias.

\subsubsection{Training and Inference} The final loss function for the KPI in training progress is as follows:
\begin{equation}
\mathcal{L} = \mathcal{L}_{\text{con}} + \lambda \cdot \mathcal{L}_{\text{sem}},
\end{equation}
where $\lambda$ denotes a tunable hyperparameter balancing contrastive discrimination and cross-modal alignment. During training, both $h_p$ and $h_{gp}$ are computed. The structure-informed $h_{gp}$ serves as a semantic teacher, guiding $h_p$ via consistency regularization. However, $h_{gp}$ depends on the patient-relevant subgraph, which in turn requires access to the ground-truth disease label. 

To ensure efficient and label-agnostic inference, $h_{gp}$ and its computation are omitted during testing. Instead, the final prediction is made by matching the learned $h_p$ against the set of disease prototypes $\{h_d\}$ using cosine similarity. This decoupled training-inference design enables KPI to benefit from graph semantics during learning while preserving fast, scalable, and interpretable prediction.

\begin{table}[]
\caption{Performance Comparison with LLMs. The best results are bold, and the second-best are underlined.}
\centering
\label{tab:llm}
\resizebox{\linewidth}{!}{
\begin{tabular}{ccccccccc}
\toprule
Metric & \multicolumn{1}{c}{LLaMA} & \multicolumn{1}{c}{GPT} & \multicolumn{1}{c}{GLM} & \multicolumn{1}{c}{Qwen} & \multicolumn{1}{c}{Claude} & \multicolumn{1}{c}{Gemini} & \multicolumn{1}{c}{DeepSeek} & \multicolumn{1}{c}{KPI} \\ \midrule
\multicolumn{9}{c}{Lung} \\ \midrule hit@1 & 0.3433 & 0.4909 & 0.6565 & 0.6158 & 0.6175 & 0.6837 & \underline{0.7519} & \textbf{0.9238} \\
hit@3 & 0.4763 & 0.7635 & 0.8510 & 0.8425 & 0.7584 & 0.8662 & \underline{0.9090} & \textbf{0.9469} \\
hit@10 & 0.5699 & 0.8198 & 0.8816 & 0.8882 & \underline{0.9550} & 0.9382 & 0.9491 & \textbf{0.9587} \\
NDCG & 0.4709 & 0.6764 & 0.8024 & 0.7662 & 0.7754 & 0.8229 & \underline{0.8652} & \textbf{0.9491} \\ \midrule
\multicolumn{9}{c}{Pneu.} \\ \midrule
hit@1 & 0.3001 & 0.4444 & 0.5906 & 0.5598 & 0.5282 & 0.6200 & \underline{0.6676} & \textbf{0.9269} \\
hit@3 & 0.4483 & 0.7276 & 0.7889 & 0.7821 & 0.6857 & \underline{0.8330} & 0.8148 & \textbf{0.9438} \\
hit@10 & 0.5547 & 0.7973 & 0.8582 & 0.8596 & 0.9140 & \underline{0.9344} & 0.8713 & \textbf{0.9512} \\
NDCG & 0.4420 & 0.6394 & 0.7591 & 0.7195 & 0.7103 & \underline{0.7889} & 0.7871 & \textbf{0.9479} \\ \midrule
\multicolumn{9}{c}{Depr.} \\ \midrule
hit@1 & 0.3353 & 0.5864 & 0.5565 & 0.6115 & 0.6371 & \underline{0.6544} & 0.6319 & \textbf{0.8119} \\
hit@3 & 0.5619 & 0.7862 & 0.7356 & \underline{0.8456} & 0.8240 & 0.8446 & 0.8418 & \textbf{0.8831} \\
hit@10 & 0.6638 & 0.8293 & 0.8141 & 0.9004 & 0.9130 & \textbf{0.9323} & \underline{0.9301} & 0.9107 \\
NDCG & 0.5124 & 0.7224 & 0.7231 & 0.7721 & 0.7826 & \underline{0.8062} & 0.7960 & \textbf{0.8812} \\ \midrule
\multicolumn{9}{c}{CHD} \\ \midrule
hit@1 & 0.2367 & 0.4651 & 0.3495 & 0.5094 & 0.5565 & 0.5770 & \underline{0.5809} & \textbf{0.6755} \\
hit@3 & 0.3826 & 0.6242 & 0.6312 & 0.6799 & 0.7382 & \textbf{0.7923} & \underline{0.7659} & 0.7168 \\
hit@10 & 0.4949 & 0.6837 & 0.8048 & 0.7548 & 0.8266 & \textbf{0.8823} & \underline{0.8427} & 0.7506 \\
NDCG & 0.3791 & 0.5828 & 0.6186 & 0.6399 & 0.6984 & \underline{0.7482} & 0.7309 & \textbf{0.7546} \\ \midrule
\multicolumn{9}{c}{Diab.} \\ \midrule
hit@1 & 0.3497 & 0.6614 & 0.3832 & 0.6888 & \underline{0.7034} & 0.6946 & 0.6847 & \textbf{0.7969} \\ 
hit@3 & 0.5378 & 0.7783 & 0.7976 & 0.8228 & \textbf{0.8773} & 0.8547 & \underline{0.8679} & 0.8284 \\
hit@10 & 0.6624 & 0.8073 & 0.9094 & 0.8469 & \underline{0.9318} & 0.9296 & \textbf{0.9568} & 0.8513 \\
NDCG & 0.5115 & 0.7408 & 0.6863 & 0.7780 & 0.8286 & 0.8222 & \underline{0.8305} & \textbf{0.8488} \\ \midrule
\multicolumn{9}{c}{Cold} \\ \midrule
hit@1 & 0.2691 & 0.4530 & 0.4986 & 0.5431 & 0.4756 & \underline{0.6166} & 0.5973 & \textbf{0.8808} \\
hit@3 & 0.4470 & 0.6841 & 0.7085 & 0.7313 & 0.6519 & \underline{0.8324} & 0.6999 & \textbf{0.8979} \\
hit@10 & 0.5838 & 0.7487 & 0.8197 & 0.8357 & 0.8681 & \textbf{0.9419} & 0.7896 & \underline{0.9186} \\
NDCG & 0.4363 & 0.6162 & 0.6906 & 0.6968 & 0.6596 & \underline{0.7914} & 0.7097 & \textbf{0.9120}\\\bottomrule
\end{tabular}
}
\end{table}

\subsection{Patient-specific Disease Explanation Generation}
In high-stakes domains like healthcare, generating reliable and interpretable explanations is essential for building trust and supporting informed decision-making. Interpretability not only helps validate predictions but also makes the model’s reasoning transparent and clinically verifiable for physicians.


KPI addresses this by generating personalized, clinically grounded justifications for its disease predictions. After producing a predicted disease label, KPI retrieves a patient-specific subgraph from the unified knowledge graph. This subgraph, together with the original patient narrative, is used to construct a prompt (illustrated in Figure~\ref{fig:prompt3}) for an LLM to generate explanations. These explanations align the model’s reasoning with domain knowledge, positioning KPI as a clinical decision-support aid that complements physician expertise with transparent and trustworthy insights.

\section{Experimental Settings}
\subsection{Baselines}
In this work, we focus on disease prediction solely from patient-side information, which is often noisy and unstructured. Unlike methods that rely on labs or imaging, this setting requires models that interpret self-reported free-text narratives. Most traditional diagnostic models are inapplicable due to their dependence on structured inputs and clinical supervision~\cite{su2024enabling,si2021deep}. We therefore compare KPI against below baselines:



\noindent\textbf{Non-medical Pre-trained Language Models (PLMs):} 
\begin{itemize}
\item \textbf{BERT}~\cite{devlin2019bert}, which is a foundational transformer-based encoder pre-trained with masked language modeling.
\item \textbf{RoBERTa}~\cite{reimers2019sentence}, which is an optimized BERT variant trained with larger corpora and dynamic masking.
\item \textbf{ALBERT}~\cite{lanalbert}, which is a parameter-efficient BERT variant leveraging factorized embeddings and weight sharing.
\item \textbf{Electra}~\cite{clarkelectra}, which is trained via replaced token detection rather than traditional MLM objectives.
\item \textbf{GIST}~\cite{solatorio2024gistembed}, which is a general semantic encoder optimized for structured sentence embeddings.
\item \textbf{Granite}~\cite{granite2024embedding}, which is a scalable PLM embedding model emphasizing dense retrieval performance.
\item \textbf{GTE}~\cite{li2023towards}, which is a dual-encoder model trained for general-purpose text embedding tasks.
\item \textbf{FinBERT}~\cite{araci2019finbert}, which is a domain-adapted BERT model pre-trained on financial texts to capture domain-specific semantics.
\end{itemize} 


\noindent\textbf{Medical PLMs:}
\begin{itemize}
\item \textbf{BioBERT}~\cite{deka2022evidence}, which is a biomedical adaptation of BERT pre-trained on PubMed abstracts.
\item \textbf{BioLORD}~\cite{remy2024biolord}, which is a contrastive biomedical encoder aligning medical concepts across modalities.
\item \textbf{MedBERT}, which is a clinically pre-trained model capturing EHR and biomedical literature semantics.
\end{itemize}
\noindent\textbf{Disease Prediction Models:}
\textbf{PoMP}~\cite{su2024enabling} is the state-of-the-art model specifically designed for patient-side disease prediction. It leverages both patient-provided textual narratives (e.g., symptoms, medical history) and basic demographic attributes (e.g., age, gender) through distinct encoders, and adopts a two-tiered classifier that first predicts broad disease categories and then refines predictions to specific diseases.

\noindent\textbf{Traditional Disease Prediction:}
We also include three machine-learning baselines: Decision Tree, KNN, and AdaBoost, which are commonly used in structured tabular data. Since these models accept only structured inputs, we evaluate them on tabular features derived from patient-provided data without raw free-text narratives.

\noindent\textbf{Large Language Models (LLMs):}
\begin{itemize}
\item \textbf{GPT-3.5 Turbo}, which is a widely used chat-optimized model by OpenAI for general-purpose reasoning.
\item \textbf{Claude 3}, which is an anthropic-developed LLM tuned for multi-turn dialogue and factual consistency.
\item \textbf{LLaMA 3.1 8B Turbo}, which is an open-source model trained with massive multilingual corpora.
\item \textbf{Gemini 1.5 Flash}, which is a Google-developed lightweight LLM with fast inference capabilities.
\item \textbf{DeepSeek V3}, which is a general LLM with strong cross-lingual performance.
\item \textbf{GLM-4 Flash}, which is a multilingual generative model with long-context handling and fast decoding.
\item \textbf{Qwen Turbo}, which is optimized for ranking and completion tasks.
\end{itemize}

Note that all PLMs are fine-tuned following the experimental setup in PoMP, and LLMs are queried using a Q\&A prompt that provides a patient narrative and asks for disease likelihood ranking. Since LLMs do not produce calibrated probability scores, AUC is not reported for their performances.

\begin{figure}[t] 
\centering 
\includegraphics[width=\linewidth]{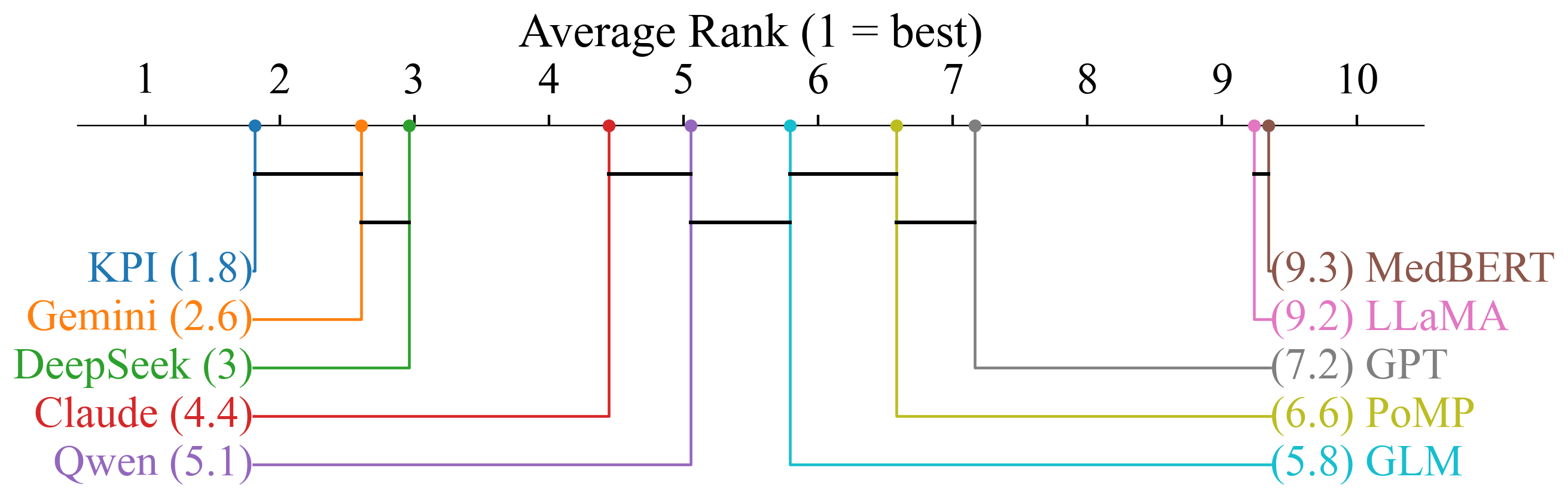} 
\caption{
Critical Difference (CD) Diagram for the top 10 models. Average ranks across datasets are compared using the Friedman test with Nemenyi post-hoc; differences are significant when indicated by p-values from the Nemenyi post-hoc at the 0.05 significance level. The x-axis shows average ranks (smaller is better; 1 at the left). Horizontal bars indicate groups with non-significant pairwise differences; within a group, the leftmost method is best ranked.}
\label{fig:cd} 
\end{figure}

\begin{figure*}[t] 
\centering 
\includegraphics[width=\linewidth]{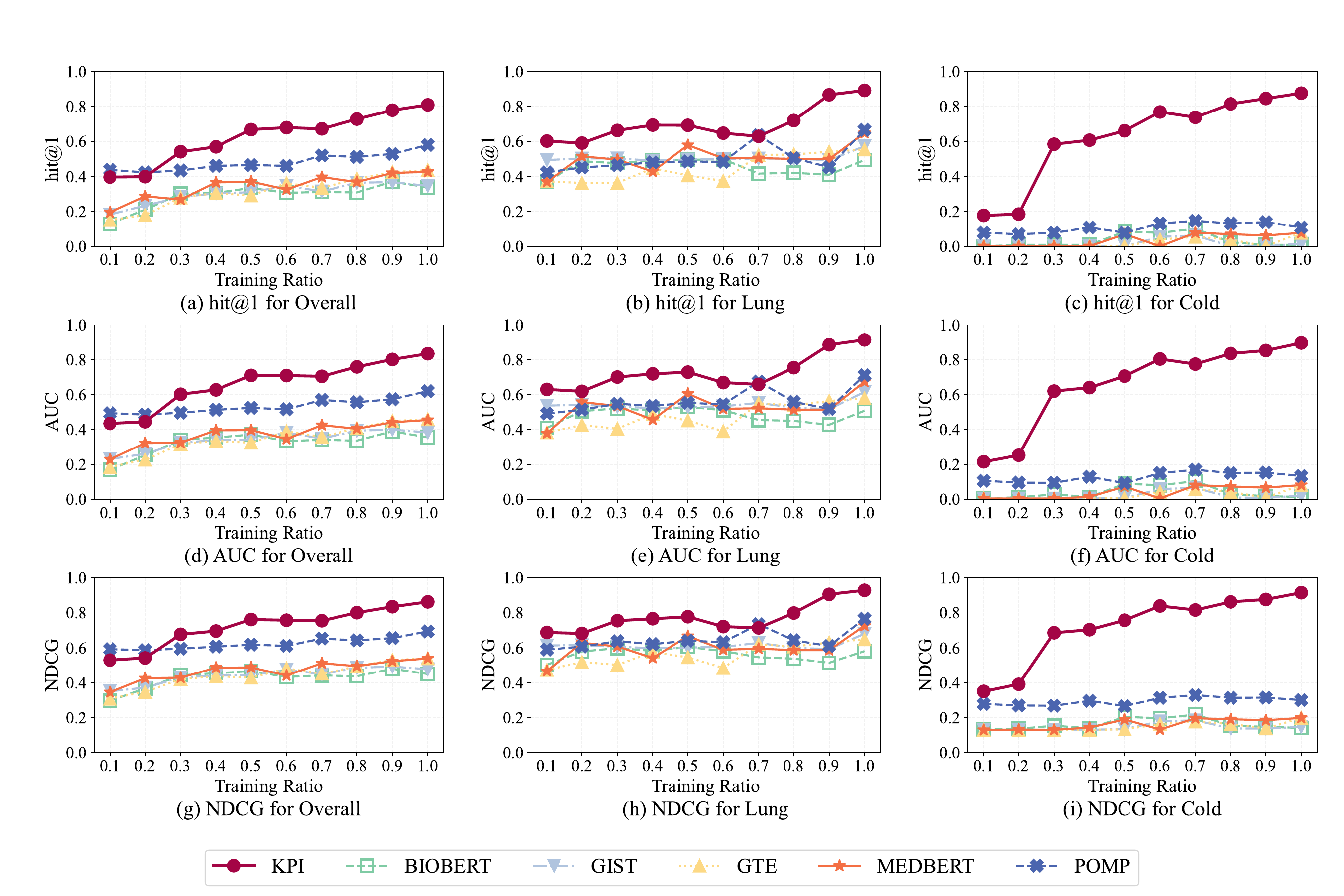} 
\caption{Scalability of KPI under Varying Training Data Sizes.} 
\label{fig:sca} 
\end{figure*}

\begin{table*}[t]
\centering
\caption{Impact of Training Data Proportion on Predictive Performance and Inference Time in the Lung Category. The best results are bold, and the second-best are underlined.}
\label{tab:time}
\resizebox{\textwidth}{!}{
\begin{tabular}{cc|ccc|ccc|ccc|ccc|ccc}
\toprule
\multicolumn{1}{l}{} & \multicolumn{1}{l|}{} & \multicolumn{3}{c|}{0\%} & \multicolumn{3}{c|}{30\%} & \multicolumn{3}{c|}{50\%} & \multicolumn{3}{c|}{70\%} & \multicolumn{3}{c}{100\%} \\
\multicolumn{1}{l}{Category} & \multicolumn{1}{l|}{Metrics} & GTE & PoMP & KPI & GTE & PoMP & KPI & GTE & PoMP & KPI & GTE & PoMP & KPI & GTE & PoMP & KPI \\ \midrule
\multirow{5}{*}{Lung} & hit@1 & 0.0000 & 0.0000 & \textbf{0.4637} & 0.0147 & \underline{0.2408} & \textbf{0.6120} & 0.1756 & \underline{0.3743} & \textbf{0.6036} & 0.3659 & \underline{0.4627} & \textbf{0.6130} & 0.5542 & \underline{0.6667} & \textbf{0.8927} \\
 & hit@3 & 0.0000 & 0.0000 & \textbf{0.6015} & 0.0421 & \underline{0.2618} & \textbf{0.6309} & 0.2072 & \underline{0.4090} & \textbf{0.6257} & 0.4206 & \underline{0.4921} & \textbf{0.6330} & 0.5657 & \underline{0.7171} & \textbf{0.9285} \\
 & hit@10 & 0.0000 & 0.0000 & \textbf{0.6320} & 0.0978 & \underline{0.2797} & \textbf{0.6593} & 0.2135 & \underline{0.4406} & \textbf{0.6383} & 0.4353 & \underline{0.5552} & \textbf{0.6519} & 0.6435 & \underline{0.8055} & \textbf{0.9401} \\
 & AUC & 0.0047 & \underline{0.0159} & \textbf{0.5426} & 0.0467 & \underline{0.2671} & \textbf{0.6309} & 0.1983 & \underline{0.4066} & \textbf{0.6207} & 0.3986 & \underline{0.4957} & \textbf{0.6290} & 0.5796 & \underline{0.7109} & \textbf{0.9137} \\
 & NDCG & 0.1293 & \underline{0.1722} & \textbf{0.6212} & 0.1803 & \underline{0.3874} & \textbf{0.6882} & 0.3083 & \underline{0.5058} & \textbf{0.6772} & 0.4876 & \underline{0.5831} & \textbf{0.6841} & 0.6490 & \underline{0.7669} & \textbf{0.9297} \\ \midrule
\multicolumn{14}{c|}{Inference Time Cost (second)} & 20.8119 & \ul{18.4767} & \textbf{17.7246}
\\\bottomrule
\end{tabular}
}
\end{table*}


\subsection{Dataset and Evaluation Metrics}
We conduct our experiments on the \textbf{Haodf} dataset, containing real-world patient-doctor consultation records covering six disease categories: Common Cold (Cold), Pneumonia (Pneu.), Diabetes (Diab.), Depression (Depr.), Coronary Heart Disease (CHD), and Lung Cancer (Lung) with statistics in Table~\ref{tab:sta}. To the best of our knowledge, \textbf{Haodf} is the only publicly available dataset that offers patient narratives paired with clinically verified diagnoses, making it a valuable benchmark for patient-side disease prediction.

To evaluate predictive quality and ranking performance, we report Hit Rate (hit@1, hit@3, and hit@10), AUC, and NDCG (without cutoff), where higher values indicate more accurate classification. Within KPI, GLM-4 Flash is employed as the LLM for knowledge-graph extraction and explanation generation, while all-MiniLM-L6-v2 is adopted for sentence embedding. For overall performance, we run each model with five random seeds and report the average.

Our goal is to develop a general model that predicts a patient’s disease based on their provided information. Note that all disease categories are considered as candidates for every test case, although evaluation is conducted separately for each category.

\section{Results and Analysis}
\subsection{Overall Performance Comparison}

To evaluate the effectiveness of KPI, we conduct comprehensive experiments across six disease categories that vary in prevalence. Table~\ref{tab:main_results} and Table~\ref{tab:llm} present the performance comparison against a broad set of baselines. We observe that KPI consistently achieves the highest hit@1 scores across all categories. 
Notably, in the category with least records, Cold, where most models perform near zero, KPI achieves a hit@1 of 0.8808 and an AUC of 0.8932, demonstrating its strong capability in handling underrepresented diseases. Moreover, in Pneumonia, tabular-based models like AdaBoost show certain advantages, possibly because of strong correlations with demographic features, underscoring the complementary value of patient-provided tabular information.
When compared with state-of-the-art LLMs, KPI consistently outperforms them on most metrics, especially hit@1, which is critical for timely clinical decisions. For CHD, KPI achieves hit@1 = 0.6755, surpassing Gemini (0.5770) and DeepSeek (0.5809), even though Gemini is slightly higher on hit@3 and hit@10.

As shown in Figure~\ref{fig:cd}, we also summarize the performances using a Critical Difference (CD) diagram over the top 10 systems. Overall differences are assessed with the Friedman test followed by Nemenyi post-hoc comparisons at $\alpha=0.05$. 
KPI attains the best average rank (1.8) and is significantly better than all baselines except Gemini and DeepSeek. Against Gemini and DeepSeek, KPI still ranks higher.
Compared with PoMP, which treats diagnosis as a purely discriminative classifier over narratives without explicit medical knowledge or prototype alignment, KPI adds structured clinical knowledge, contrastive disease prototypes, and interpretable subgraphs, leading to better calibration and generalization, especially for long-tailed diseases. Different from general-purpose LLMs, KPI’s domain constraints and ranking-oriented objective reduce hallucination and capture disease-specific cues more reliably.

\begin{table}[]
\caption{Ablation study results evaluating the impact of data formulation and architectural components on KPI’s performance across six disease categories. The best results are bold, and the second-best are underlined.
}
\centering
\label{tab:abl}
\resizebox{\linewidth}{!}{
\begin{tabular}{ccccccc}
\toprule
\multicolumn{1}{l}{Category} & \multicolumn{1}{l}{Metrics} & Textual & Textual w/ S & KPI w/o S.C & KPI w/o P.K & KPI \\ \midrule
\multirow{5}{*}{Lung} & hit@1 & 0.7802 & 0.7666 & \ul{0.8780} & 0.7981 & \textbf{0.8927} \\
 & hit@3 & 0.8023 & 0.7802 & \ul{0.9169} & 0.8055 & \textbf{0.9285} \\
 & hit@10 & 0.8055 & 0.7876 & \ul{0.9317} & 0.8170 & \textbf{0.9401} \\
 & AUC & 0.7943 & 0.7765 & \ul{0.8994} & 0.8054 & \textbf{0.9137} \\
 & NDCG & 0.8274 & 0.8090 & \ul{0.9180} & 0.8343 & \textbf{0.9297} \\ \midrule
\multirow{5}{*}{Pneu.} & hit@1 & 0.7652 & 0.7971 & \ul{0.8930} & 0.7987 & \textbf{0.8962} \\
 & hit@3 & 0.7812 & 0.8067 & \ul{0.9026} & 0.8035 & \textbf{0.9217} \\
 & hit@10 & 0.8067 & 0.8131 & \ul{0.9169} & 0.8131 & \textbf{0.9297} \\
 & AUC & 0.7810 & 0.8063 & \ul{0.9014} & 0.8049 & \textbf{0.9116} \\
 & NDCG & 0.8182 & 0.8366 & \ul{0.9182} & 0.8331 & \textbf{0.9270} \\ \midrule
\multirow{5}{*}{Depr.} & hit@1 & 0.7198 & 0.7117 & 0.7485 & \textbf{0.7935} & \ul{0.7853} \\
 & hit@3 & 0.7464 & 0.7710 & 0.7935 & \ul{0.7975} & \textbf{0.8569} \\
 & hit@10 & 0.7853 & 0.7996 & \ul{0.8262} & 0.8139 & \textbf{0.8793} \\
 & AUC & 0.7423 & 0.7487 & 0.7793 & \ul{0.8017} & \textbf{0.8274} \\
 & NDCG & 0.7850 & 0.7924 & 0.8180 & \ul{0.8330} & \textbf{0.8598} \\ \midrule
\multirow{5}{*}{CHD} & hit@1 & 0.4480 & 0.4276 & 0.5385 & \textbf{0.5973} & \ul{0.5701} \\
 & hit@3 & 0.4706 & 0.4457 & 0.5769 & \textbf{0.6109} & {\ul0.6018} \\
 & hit@10 & 0.4932 & 0.4661 & \ul{0.6244} & 0.6222 & \textbf{0.6448} \\
 & AUC & 0.4718 & 0.4501 & 0.5726 & \textbf{0.6096} & \ul{0.5985} \\
 & NDCG & 0.5558 & 0.5375 & 0.6442 & \textbf{0.6655} & \ul{0.6644} \\ \midrule
\multirow{5}{*}{Diab.} & hit@1 & 0.6682 & \ul{0.6751} & 0.6110 & 0.6293 & \textbf{0.7574} \\
 & hit@3 & 0.6773 & \ul{0.6796} & 0.6453 & 0.6522 & \textbf{0.7849} \\
 & hit@10 & 0.6819 & \ul{0.6865} & 0.6636 & 0.6590 & \textbf{0.8238} \\
 & AUC & 0.6791 & \ul{0.6826} & 0.6338 & 0.6467 & \textbf{0.7794} \\
 & NDCG & 0.7290 & \ul{0.7298} & 0.6915 & 0.7001 & \textbf{0.8168} \\ \midrule
\multirow{5}{*}{Cold} & hit@1 & 0.6615 & 0.7077 & 0.8462 & \ul{0.8615} & \textbf{0.8769} \\
 & hit@3 & 0.6692 & 0.7154 & 0.8538 & \ul{0.8692} & \textbf{0.9077} \\
 & hit@10 & 0.6846 & 0.7308 & \ul{0.8769} & \ul{0.8769} & \textbf{0.9231} \\
 & AUC & 0.6768 & 0.7177 & 0.8574 & \ul{0.8695} & \textbf{0.8959} \\
 & NDCG & 0.7314 & 0.7619 & 0.8823 & \ul{0.8903} & \textbf{0.9158}\\ \bottomrule
\end{tabular}
}
\end{table}



\subsection{Accuracy and Inference Efficiency under Data Imbalance}

We evaluate KPI’s scalability and robustness by testing long-tailed disease prediction, data imbalance, and inference efficiency. In practice, many categories have sparse annotations and highly skewed distributions, so models should generalize under limited supervision and run efficiently at scale.

First, we evaluate KPI in low-resource settings by subsampling the training data from 100\% down to 10\% and comparing it with the top five baselines. As shown in Figure~\ref{fig:sca}, KPI outperforms all baselines at nearly all training ratios. With 10–20\% data, KPI and PoMP are comparable, likely because some categories (e.g., Lung), still provide enough samples for PoMP to leverage PLM priors. KPI leads on long-tail category Cold. With only 30\% data, KPI makes a clear jump and reaches performance competitive with full-data training.


Second, unlike the prior experiment that subsampled all classes, here we subsample only the Lung cases and keep all others (and the test set) fixed. Shown in Table~\ref{tab:time}, we vary cases of Lung from 0\% to 100\%, simulating a long-tail setting. We observe that KPI remains strong across all proportions, rises quickly with more Lung data, stabilizes between 30\% and 70\%, and with full data reaches hit@1 = 0.8927 and AUC = 0.9137. GTE and PoMP lag throughout, indicating KPI’s robustness to imbalance data.


Beyond accuracy, we also assess inference efficiency in Table~\ref{tab:time}. KPI completes inference on the whole test set in 17.72 seconds, outperforming the top baselines PoMP and GTE. This demonstrates that KPI is both effective and practical for real-world deployment.

\begin{figure}[t] 
\centering 
\includegraphics[width=0.9\linewidth]{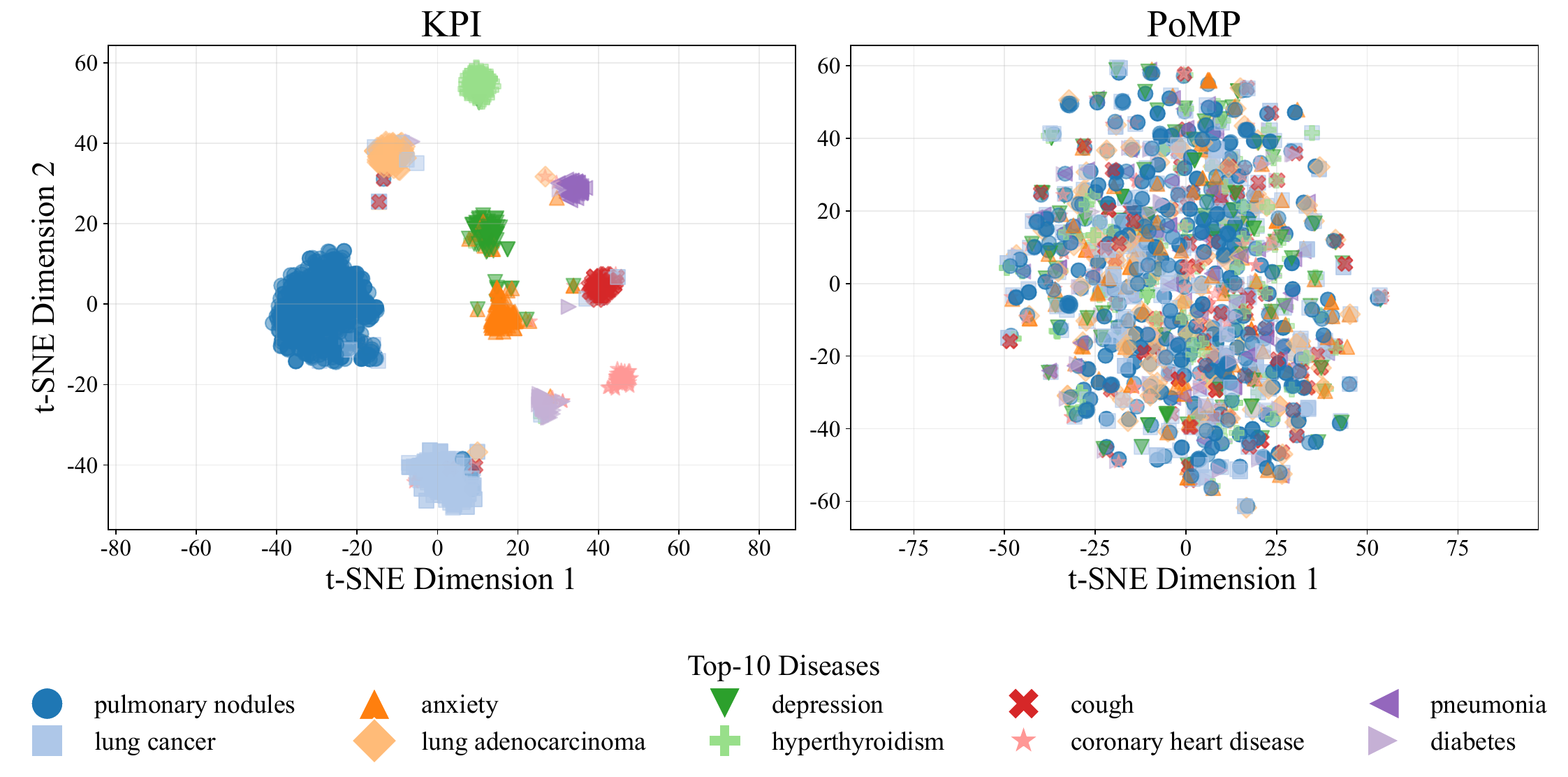} 
\caption{Patient Embeddings Visualization via T-SNE: KPI vs. PoMP on the Top 10 Diseases (Color-Coded).}
\label{fig:tsne} 
\end{figure}

\begin{figure}[t] 
\centering 
\includegraphics[width=0.7\linewidth]{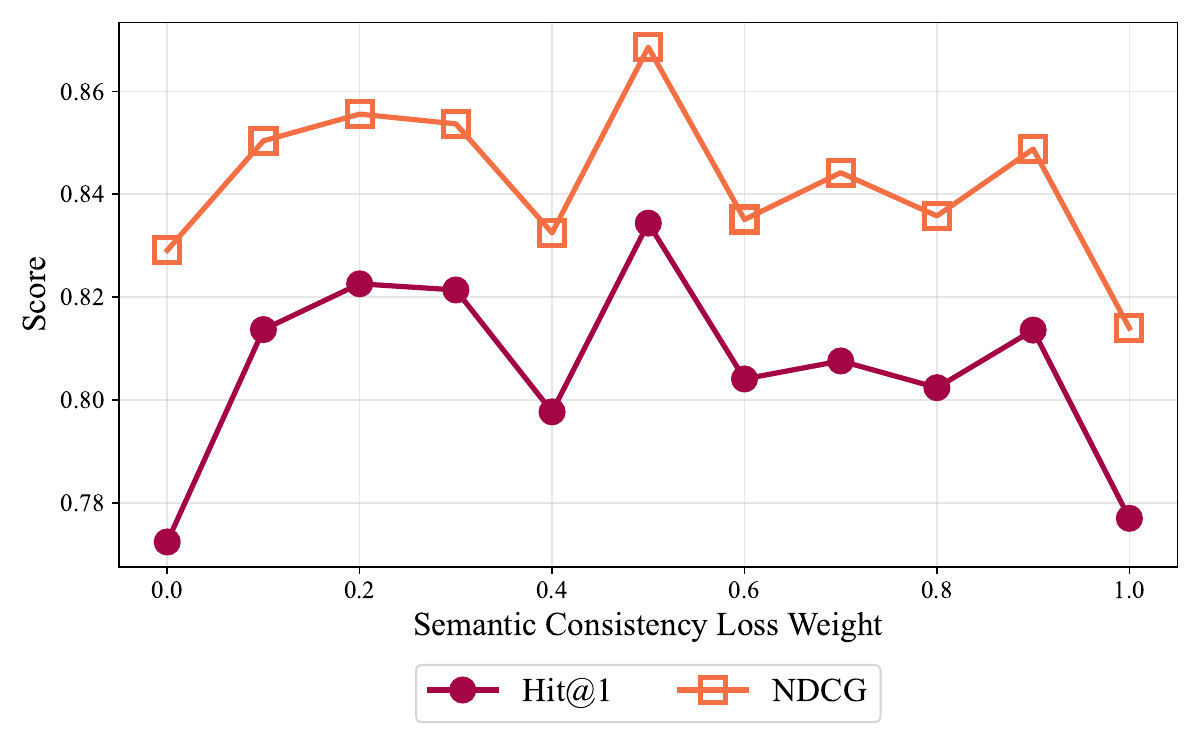} 
\caption{Effect of Varying Semantic Consistency Loss Weight ($\lambda$) on the Overall Performance Across All Categories.} 
\label{fig:cl} 
\end{figure}

\begin{table*}[t]
  \caption{Case Study}
  \resizebox{\textwidth}{!}{
  \centering
  \label{tab:case}
  \renewcommand{\arraystretch}{1.2}
  \begin{tabular}{
    >{\arraybackslash}m{0.18\linewidth} 
    >{\centering\arraybackslash}m{0.3\linewidth} 
    >{\arraybackslash}m{0.22\linewidth}
    >{\arraybackslash}m{0.12\linewidth}
    >{\arraybackslash}m{0.17\linewidth}
  }
    \toprule
    \textbf{Patient Narrative} & 
    \textbf{KPI's Patient-relevant Subgraph} & 
    \textbf{KPI's Output} &
    \textbf{PoMP's Output} &
    \textbf{Gemini's Output} \\
    \midrule
    I have a stuffy nose, \textcolor{Red}{\textbf{cough, and sore throat}}. I had a \textcolor{Green}{\textbf{fever}} for the past two days. Now I have taken medicine to reduce the fever. My \textcolor{Red}{\textbf{throat is hoarse}}. It hurts so much when I cough. Now I feel like there is \textcolor{Blue}{\textbf{phlegm in my respiratory tract}} and I can’t cough it out... 
    & 
    \includegraphics[width=\linewidth]{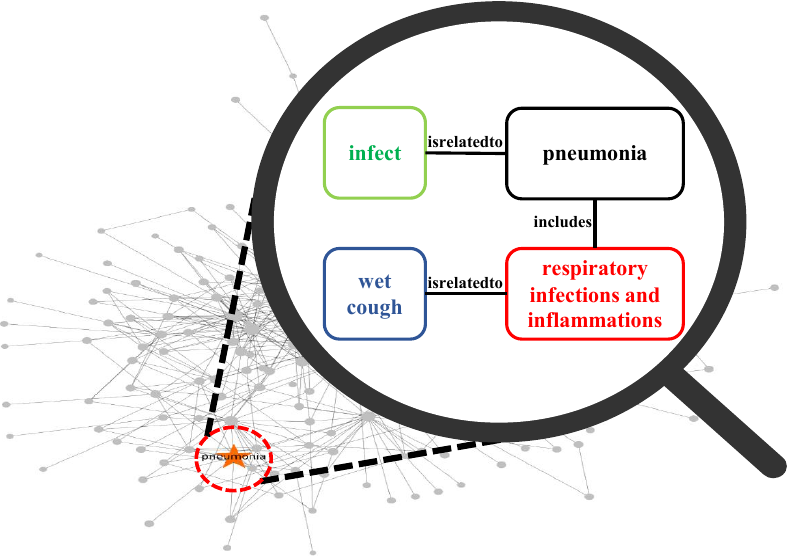}
    & 
    \textbf{Predicted Disease}: pneumonia. \textbf{Explanation}: The patient's symptoms—\textcolor{Green}{\textbf{fever}}, \textcolor{Red}{\textbf{cough, sore throat}}, and \textcolor{Blue}{\textbf{phlegm}}—align with pneumonia, a lung infection. High attention score triples link pneumonia to \textcolor{Red}{\textbf{respiratory infections}} and chronic lung diseases like COPD, supporting this diagnosis. 
    &
    \textbf{Predicted Disease}: congenital heart disease (0.0273). \textbf{Explanation}: None.
    &
    \textbf{Predicted Disease}: cold. \textbf{Explanation}:  The patient reports symptoms like a stuffy nose, cough, sore throat, and fever, which are characteristic of a common cold.
    \\
    \bottomrule
  \end{tabular}
  }
\end{table*}

\subsection{Ablation Study}

\subsubsection{Contribution of KPI Components}
To better understand the contributions of each component in KPI, we run ablations on data formulation and architecture design. Table VI summarizes the performance of several variants across six diseases. 

From data side, \textbf{Textual} uses only patient narratives, while \textbf{Textual w/ S} adds self-reported attributes (e.g., age) converted to natural language; unlike full KPI, both treat all inputs as text. From architecture side, \textbf{KPI w/o S.C} removes semantic consistency loss, and \textbf{KPI w/o P.K} replaces knowledge-derived prototypes with randomly initialized embeddings.


Experimental results show that removing either component reduces model accuracy, especially in low-resource categories such as Cold and Coronary Heart Disease. For example, dropping knowledge in prototype construction reduces AUC from 0.9137 to 0.8054 on Lung. Likewise, the data-only variants perform worse than the full model, highlighting the benefit of combining structured and unstructured inputs. Notably, in Coronary Heart Disease, \textbf{KPI w/o P.K.} is slightly better than KPI, likely because standardized diagnostic criteria make narratives and basic attributes highly informative. Overall, KPI outperforms its variants in most categories and metrics, underscoring the value of both data integration and knowledge-guided design in patient-side disease prediction.

\subsubsection{Representation Analysis}

To gain deeper insights into the learned patient representations, we project text-based patient embeddings into a 2D space using t-SNE. As shown in Figure~\ref{fig:tsne}, patients with the same disease labels form clear and well-separated clusters in KPI, whereas the patient embeddings are largely intermixed in PoMP. This indicates that KPI can capture semantically meaningful structures more effectively.

\subsubsection{Effect of Semantic Consistency Loss}
We tune the semantic consistency weight $\lambda$ and find $\lambda = 0.5$ gives the best overall performance (Figure~\ref{fig:cl}), with similar gains in Lung and Cold. Performance drops when $\lambda$ is too low (insufficient grounding) or too high (over-regularization), indicating that semantic consistency works best when properly balanced.

\subsection{Explanation Analysis}
To evaluate the quality of the explanations generated by KPI, we conduct a systematic assessment. Interpretability is crucial for building clinician trust and supporting effective patient communication, we aim to evalueate whether KPI’s explanations are both medically accurate and contextually relevant to individual patients. 
Specifically, we randomly select 60 cases, covering all six disease categories. 
Each explanation is then reviewed by four licensed physicians with relevant clinical experience. The evaluation is conducted from three complementary perspectives: 1) \textit{Medical Soundness}: Whether the explanation accurately reflects clinical characteristics, pathophysiological mechanisms, or diagnostic reasoning related to the disease; 2) \textit{Patient Alignment}: Whether the explanation is tailored to the patient's narrative, such as symptoms, disease course, or physical findings; and 3) \textit{Clarity and Coherence}: Whether the explanation is well-structured, logically coherent, and easy to understand.
Each criterion is rated on a scale from 0 to 5, with higher scores indicating better quality. 

On average, the explanations received high expert ratings: \textbf{4.3} for \textit{Medical Soundness}, \textbf{4.1} for \textit{Patient Alignment}, and \textbf{4.2} for \textit{Clarity and Coherence}. These results indicate that KPI is capable of generating explanations that are clinically accurate, well-aligned with patient narratives, and clearly articulated, highlighting its potential for real-world interpretability.

To further illustrate KPI’s effectiveness, Table~\ref{tab:case} shows a representative case. KPI extracts a patient-specific subgraph from the KG and grounds its explanation in structured relations, such as mapping “phlegm in my respiratory tract” to wet cough. In contrast, PoMP returns only a label and probability, and Gemini produces a general explanation without evidence from reliable sources. By leveraging knowledge-guided subgraphs, KPI provides explanations that are semantically reliable and clinically coherent.




\section{Conclusion}
In this work, we propose KPI, a unified framework for disease prediction based solely on patient-side information. By integrating structured medical knowledge, contrastive prototype learning, and LLM-based explanation generation, KPI addresses key challenges in handling unstructured narratives, long-tailed diseases, and lack of interpretability. Experiments on real-world datasets show that KPI outperforms existing baselines in both accuracy and explanation quality. 
Designed as decision support, we believe that KPI can serve as a practical assistive triage tool that complements clinical workflows under physician supervision. As future work, we will analyze KPI failure cases and investigate strategies to extend inputs beyond text to multimodal signals, and scale to broader disease domains while preserving interpretability and efficiency.



\section*{Acknowledgment}
This research is supported, in part, by (1) the Joint NTU-UBC Research Centre of Excellence in Active Living for the Elderly (LILY), Nanyang Technological University, Singapore; and (2) Jinan-NTU Green Technology Research Institute (GreenTRI).

\section*{AI-Generated Content Acknowledgement}
Portions of this paper are proofread and refined for spelling and grammatical accuracy using ChatGPT.

\bibliographystyle{IEEEtran}
\bibliography{ref}

\end{document}